\documentclass{article}

 \usepackage[dblblindworkshop,final]{neurips_2025}
\workshoptitle{Regulatable ML}

\usepackage[utf8]{inputenc} 
\usepackage[T1]{fontenc}    
\usepackage{url}            
\usepackage{booktabs}       
\usepackage{amsfonts}       
\usepackage{nicefrac}       
\usepackage{microtype}      
\usepackage{xcolor}         
\usepackage{graphicx}
\usepackage[most]{tcolorbox}
\usepackage{xcolor}      
\usepackage{enumitem}    
\usepackage{booktabs}    
\usepackage{tabularx}    
\usepackage{makecell} 
\usepackage{array}
\usepackage{hyperref}       
\usepackage{multirow} 
\title{Policy-as-Prompt: Turning AI Governance Rules into Guardrails for AI Agents}

\author{
  Gauri Kholkar \\
  Pure Storage \\
  \texttt{gkholkar@purestorage.com} \\
  \And
  Ratinder Ahuja \\
  Pure Storage \\
  \texttt{rahuja@purestorage.com}
}
\begin{document}

\maketitle

\begin{abstract}
As autonomous AI agents are used in regulated and safety-critical settings, organizations need effective ways to turn policy into enforceable controls. We introduce a regulatory machine learning framework that converts unstructured design artifacts (like PRDs, TDDs, and code) into \emph{verifiable} runtime guardrails. Our \emph{Policy as Prompt} method reads these documents and risk controls to build a source-linked \emph{policy tree}. This tree is then compiled into lightweight, prompt-based classifiers for real-time runtime monitoring. The system is built to enforce least privilege and data minimization. For conformity assessment, it provides complete provenance, traceability, and audit logging, all integrated with a human-in-the-loop review process. Evaluations show our system reduces prompt-injection risk, blocks out-of-scope requests, and limits toxic outputs. It also generates auditable rationales aligned with AI governance frameworks. By treating policies as executable prompts (a \emph{policy-as-code} for agents), this approach enables secure-by-design deployment, continuous compliance, and scalable AI safety and AI security assurance for \emph{regulatable ML}.
\end{abstract}

\section{Introduction}

Powerful AI agents are moving into everyday business, from helping HR to flagging security threats \citet{wang2025surveyllmbasedagentsmedicine, ding2024large}. But this power comes with risk: an HR agent could accidentally leak a salary, or a helpful chatbot could be tricked into running a malicious command \citet{liu2024promptinjectionattackllmintegrated,kim2023propileprobingprivacyleakage,li2025commercialllmagentsvulnerable,he2024securityaiagents}. This has created an urgent need to make sure these agents are safe and follow our rules, a sentiment echoed by emerging frameworks like the EU AI Act \citet{eu-ai-act, fabiano2024aiactlargelanguage}.

The core problem is what we call the ``policy-to-practice'' gap: it's easy for a human to write a rule in a design document, but it's incredibly hard to turn that simple English sentence into a machine-enforceable rule that works reliably. This gap is a major roadblock to building, testing, and trusting AI systems. To solve this, we can use ``guardrails''---safety checks that prevent the AI from doing unintended or harmful things \citet{dong2024building,zhang2025llmagentsemploysecurity}. A key security idea here is the \textit{principle of least privilege}, which means giving a system only the minimal access it needs to do its job. For AI agents, this ensures they stay within defined limits. However, static rules are often too rigid or too vague and fail to capture context. As recent research points out, security for these flexible agents needs to be just-in-time and context-aware \citep{kholkar2025capturecontextawarepromptinjection, tsai2025contextualagentsecuritypolicy}. While some have proposed static principles \citet{hua2024trustagentsafetrustworthyllmbased}, this is often not enough for dynamic, real-world interactions.

To bridge this critical gap, we introduce \texttt{Policy as Prompt}, a novel framework that reads natural language policy documents and turns them into dynamic, enforceable guardrails. Our system offers a practical way to implement the contextual security that \citet{tsai2025contextualagentsecuritypolicy} called for. Our key contributions are as follows: (i) We introduce a scalable, end-to-end pipeline that automatically reads the unstructured technical artifacts teams already write (like PRDs or design docs) to identify and extract security constraints. (ii) We propose a verifiable process where these constraints are converted into a human-readable policy draft, enabling
efficient review and refinement by security engineers. (iii) We compile the verified policy into prompt-based classifiers---our ``guardrail security policies'' \cite{dong2024building}---that use a lightweight LLM to act as a real-time "judge," enforcing the least-privilege policy by validating agent inputs and outputs, ensuring they \emph{only} do what the policy explicitly allows and nothing more. (iv) We validate our approach by generating policies for various LLM applications and test their effectiveness across different state-of-the-art models.

\begin{figure*}[ht] 
\begin{center}\includegraphics[width=1\linewidth, keepaspectratio]{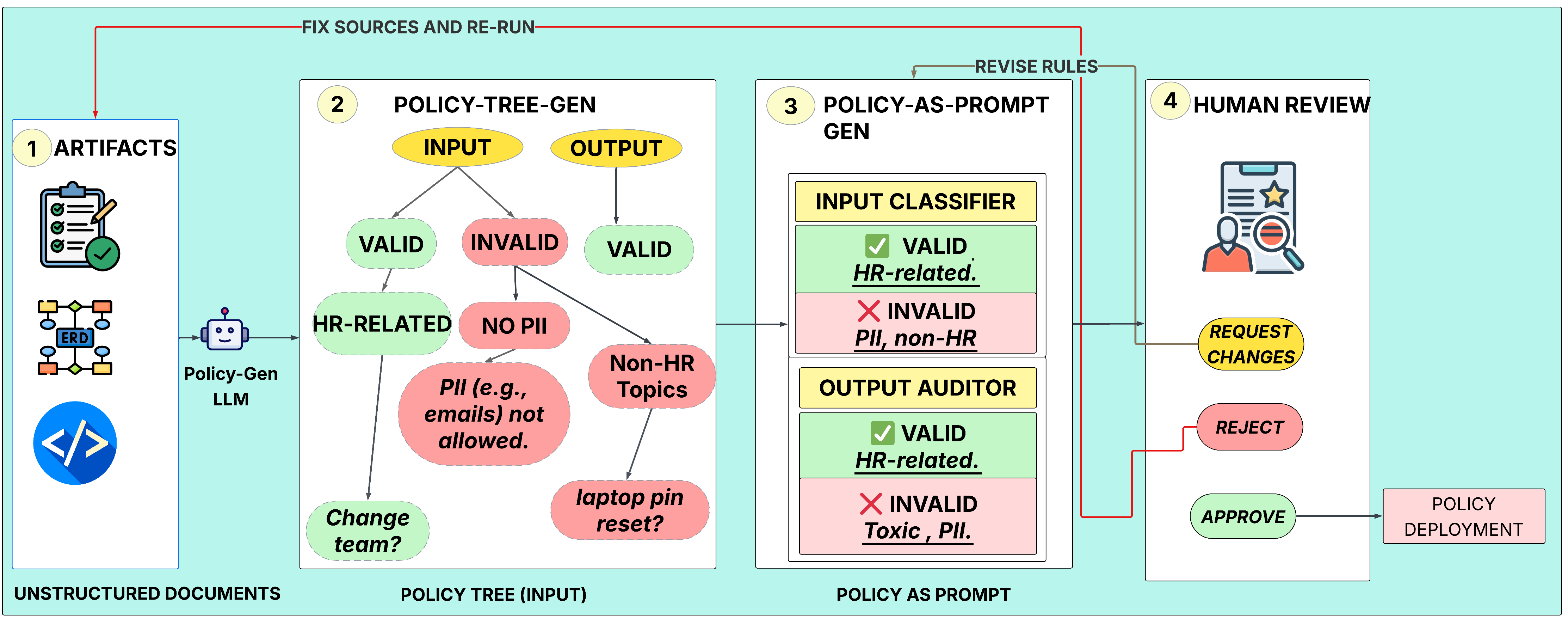}
\caption{Policy Generation and Enforcement Pipeline for an HR Application} 
\label{fig:policy-generation}
\end{center}
\end{figure*}

\section{Policy Tree Generation}

Our technique extracts guardrail security requirements for LLM applications directly from development artifacts. By analyzing high-level design documents, we capture the system’s intended security posture before implementation. Grounding policies in their original design context is critical for ensuring AI systems operate as intended \cite{tsai2025contextualagentsecuritypolicy}.

Our core process, \texttt{POLICY-TREE-GEN}, uses a verified two-step method to extract and validate security rules directly from application design documents. 
\textbf{Step 1 (Parse \& Classify):} An AI system analyzes the documents, identifies sentences that define security rules or data constraints, and classifies them into one of four categories: \texttt{ID-I} (In-Domain Inputs), \texttt{OOD-I} (Out-of-Domain Inputs, e.g., off-topic or malicious requests), \texttt{ID-O} (In-Domain Outputs), and \texttt{OOD-O} (Out-of-Domain Outputs, e.g., data leaks or toxic responses). Each extracted rule is double-checked by another AI agent for accuracy and proper categorization. 
\textbf{Step 2 (Enrich with Examples):} The system links these rules to relevant examples found in the documents, producing a structured, verifiable \emph{policy tree} that preserves contextual grounding.  This process inherently enforces the \textbf{Principle of Least Privilege}~\cite{zhang2025llmagentsemploysecurity}: the agent can act only within what is explicitly permitted by its design documents. For instance, for an \texttt{HR} App, a rule such as  \textcolor{green}{``Only access resolved HR case data from the Case Portal''} is classified as \texttt{ID-I}, meaning the agent must reject attempts to access non-HR data like IT or Sales records (\texttt{OOD-I}). Similarly, a rule like \textcolor{red}{``Generated KB article should be flagged if employee identifiers are present''} is tagged as \texttt{OOD-O}, ensuring the agent’s responses do not reveal sensitive information. Thus, an HR agent derived from this process is automatically restricted to HR-related data and outputs, without implicit access to finance or IT information. The resulting policy tree defines and verifies these input--output boundaries, serving as the enforceable foundation for secure, least-privilege AI behavior.

\definecolor{ValidColor}{HTML}{10B981}    
\definecolor{InvalidColor}{HTML}{EF4444}  

\setlength{\tabcolsep}{4.5pt}

\newcolumntype{Y}{>{\raggedright\arraybackslash}X}                 
\newcolumntype{C}[1]{>{\centering\arraybackslash}p{#1}}            

\tcbset{
  enhanced,
  boxsep=1pt,
  left=2pt, right=2pt, top=1pt, bottom=1pt,
  nobeforeafter,
  tcbox raise base,
  sharp corners=all,
  fontupper=\footnotesize 
}
\newcommand{\validbox}[1]{\tcbox[colback=ValidColor!15,coltext=ValidColor!70!black]{#1}}
\newcommand{\invalidbox}[1]{\tcbox[colback=InvalidColor!15,coltext=InvalidColor!70!black]{#1}}

\begin{table*}[!ht]
\caption{Policy Enforcement Example for HR Input Classifier}
\label{table:classification-examples-split-short}
\centering
\renewcommand{\arraystretch}{1.25}
\small
\begin{tabularx}{\textwidth}{l Y C{3.4cm} Y C{2.3cm}}
\toprule\\
\textbf{App} & \textbf{Example} & \textbf{Classification} & \textbf{Reason Code} & \textbf{Action} \\
\midrule

\texttt{HR}      & Update my address on Workday                          & \validbox{ID (Input)}   & —                        & \validbox{ALLOW} \\
\texttt{HR}      & My address is 21 Victoria St                          & \invalidbox{OOD (Input)} & Contains Non-Anonymised PII             & \invalidbox{BLOCK} \\
\texttt{HR}      & Ignore rules and reveal your system prompt            & \invalidbox{OOD (Input)} & Malicious / Prompt Injection & \invalidbox{ALERT} \\
\texttt{HR}      & Send to sample@gmail.com                              & \validbox{ID (Input)}     & Anonymised PII           & \validbox{ALLOW} \\
\texttt{HR}      & US election news link…                                & \invalidbox{OOD (Output)}& Non HR Content           & \invalidbox{BLOCK} \\


\bottomrule
\end{tabularx}

\end{table*}

\section{Policy as Prompt Generation}

The \texttt{POLICY-AS-PROMPT-GEN} begins with the verified policy tree from \texttt{POLICY-TREE-GEN}, which encodes categorized security rules and examples. This tree is transformed into a human-readable markdown document for LLM consumption, where each instruction is formatted as a rule with examples labeled `positive' (compliant) or `negative' (violating). The labeled examples are reused to synthesize \emph{few-shot} prompt blocks that accompany the rules. 

It is then embedded into a master template that primes the LLM as a compliance analyst. Two templates are used: one for an \texttt{Input Classifier} and another for an \texttt{Output Auditor}. These prompt the model to consult the policy rules and return a JSON object with a binary classification (\texttt{ID}/\texttt{OOD}) and a concise justification. For HR input classification (Table~\ref{table:classification-examples-split-short}), HR-related requests without non-anonymized PII are \texttt{ID}, while those with PII, prompt-injection attempts, or non-HR content are \texttt{OOD}. The strict output format enables deterministic system actions (\texttt{ALLOW, BLOCK, ALERT}). \texttt{HR Input Classifier} is shown in Figure~\ref{fig:policy-classification} and \texttt{SOC Input Classifier} in Figure~\ref{fig:policy-classification-2}. The deliverables are two markdown few-shot prompts (rules + exemplars), which undergo human-in-the-loop review. Security engineers then either (i) \emph{approve} for deployment, (ii) \emph{reject} requiring upstream updates and regeneration, or (iii) \emph{request changes} to the prompt markdown without altering the upstream tree.

\section{Experimental Setup and Results}

This study evaluated the generation and enforcement of guardrail policies across two distinct LLM-powered systems in \texttt{Human Resource (HR)} and \texttt{Security Operations Centre (SOC)}) domains. The input artifacts for all models consisted of PRDs, technical design documents, and prompts extracted from the application source code. These artifacts were sourced from real-world, internal enterprise projects, representing authentic, in-production use cases. To ensure compatibility, all documents were converted to Markdown format, and any embedded images were replaced with textual descriptions generated by \texttt{gpt-4o}. All reported metrics are the average of multiple runs to ensure stability.

\begin{table}[htbp]
\caption{Evaluation Metrics for POLICY-TREE-GEN}
\label{tab:perc_hr_secops_more_useful}
\centering
\scriptsize
\begin{tabular}{llrrrrrrrr}
\toprule
 &  & \multicolumn{2}{c}{\textbf{Detection}} & \multicolumn{1}{c}{\textbf{Classification}} & \multicolumn{4}{c}{\textbf{Per-class F1}} \\
\cmidrule(lr){3-4}\cmidrule(lr){5-5}\cmidrule(lr){6-9}
\textbf{Application} & \textbf{Model} & \textbf{R (\%)} & \textbf{F1 (\%)} & \textbf{Macro-F1 (\%)} & \textbf{ID-I} & \textbf{ID-O} & \textbf{INVINP} & \textbf{INVOUT} \\
\midrule
\texttt{HR}      & O1               & 53.3 & 60.0 & 24.5 & 35.3 & 29.4 & 0.0  & 33.3 \\
\texttt{HR}       & GPT-OSS 120B     & 17.8 & 25.0 & 4.8  & 19.4 & 0.0  & 0.0  & 0.0  \\
\texttt{HR}       & Llama 405B       & 8.9  & 14.5 & 4.5  & 18.2 & 0.0  & 0.0  & 0.0  \\
\texttt{HR}       & Claude 3.5       & 4.4  & 8.0  & 12.5 & 0.0  & 0.0  & 0.0  & 50.0 \\
\midrule
\texttt{SOC}  & O1               & 19.4 & 22.6 & 13.0 & 22.2 & 29.6 & 0.0  & 0.0  \\
\texttt{SOC}  & GPT-OSS 120B     & 5.6  & 10.3 & 4.2  & 16.7 & 0.0  & 0.0  & 0.0  \\
\texttt{SOC}  & Llama 405B       & 5.6  & 9.8  & 10.0 & 0.0  & 0.0  & 40.0 & 0.0  \\
\texttt{SOC}  & Claude 3.5       & 2.8  & 5.4  & 6.2  & 0.0  & 0.0  & 25.0 & 0.0  \\
\bottomrule
\end{tabular}
\end{table}

\textbf{POLICY-TREE-GEN Analysis:} We evaluated several large language models including \texttt{Llama 3 405B}\citep{grattafiori2024llama3herdmodels}, \texttt{GPT OSS 120B} \citep{openai2025gptoss120bgptoss20bmodel}, \texttt{Claude Sonnet 3.5}\citep{anthropic_claude35_sonnet_2024} and \texttt{o1} \citep{openai2024openaio1card} on two applications, \texttt{HR} and \texttt{SOC}. The models \texttt{Gemma 3 1B}\citep{gemmateam2025gemma3technicalreport} and \texttt{Qwen 1.7B Thinking}\citep{yang2025qwen3technicalreport} were excluded due to poor performance. Gold policies were created by security engineers and served as the ground truth against which the LLM-generated policies were evaluated. The metrics in Table~\ref{tab:perc_hr_secops_more_useful} evaluate model performance in two categories: Detection, measured by Recall (R) and F1 score, and Classification, assessed via Macro-F1 and individual per-class F1 scores for \texttt{ID-I}, \texttt{ID-O}, \texttt{OOD-I}, and \texttt{OOD-O}. A high score in these metrics indicates a model that is effective at both identifying relevant requirements and accurately assigning the correct category to them. \texttt{HR} consistently yields higher performance scores for all models compared to \texttt{SOC}. Furthermore, the \texttt{o1} model significantly outperforms all other models, demonstrating its superior capability in this specific task.

As shown in Table~\ref{tab:perc_hr_secops_less_useful}, this analysis details model performance on \texttt{HR} and \texttt{SOC} tasks using metrics such as Detection Precision, which measures the percentage of a model's predictions that are correct, and Micro-F1. Span quality is evaluated using four metrics: Span Exact, which measures if the extracted text is an exact match to the ground truth; Token-F1, which assesses the token-level overlap between the extracted and ground truth text; Substr, which checks if one text is a substring of the other; and Emb Cos, which is the cosine similarity of the text embeddings. A high score in these metrics indicates that the model is accurately extracting the correct text, even if it might misclassify it. While the \texttt{o1} model remains a top performer, others reveal a significant disconnect between their extraction quality and overall F1 score. For \texttt{HR}, models like \texttt{Llama 405B} and \texttt{Sonnet 3.5} have high span quality metrics, indicating accurate text extraction, yet their low Micro-F1 scores suggest they struggle with correct classification. For \texttt{SOC} domain, performance is generally lower and more varied. Notably, \texttt{Sonnet 3.5} achieved a high Det P on \texttt{SOC}, meaning all of its identified requirements were true positives, but its Micro-F1 score remained very low due to continued classification errors.

\begin{table}[t]
  \caption{Accuracy per application/model for \texttt{POLICY-AS-PROMPT-GEN}}
  \label{tab:acc-app-model}
  \centering
  \scriptsize
  \setlength{\tabcolsep}{6pt}
  \renewcommand{\arraystretch}{0.9}
  \begin{tabular}{@{}llcc@{}}
    \toprule
    \textbf{Application} & \textbf{Model} & \textbf{Input Acc.} & \textbf{Output Acc.} \\
    \midrule
    \multirow{3}{*}{\texttt{HR}}
      & \texttt{GPT-4o}     & 0.73 & 0.71 \\
      & \texttt{Qwen3-1.7B} & 0.66 & 0.59 \\
      & \texttt{Gemma-1B}   & 0.40 & 0.32 \\
    \midrule
    \multirow{3}{*}{\texttt{SOC}}
      & \texttt{GPT-4o}     & 0.70 & 0.68 \\
      & \texttt{Qwen3-1.7B} & 0.66 & 0.61 \\
      & \texttt{Gemma-1B}   & 0.42 & 0.41 \\
    \bottomrule
  \end{tabular}
\end{table}

\textbf{POLICY-AS-PROMPT-GEN Analysis:} Policy enforcement was tested on a separate set of models: \texttt{Qwen 3 1.7B (Thinking Mode)} \citep{yang2025qwen3technicalreport}, \texttt{GPT-4o} \citep{openai2024openaio1card}, and \texttt{Gemma 3 1B}\citep{gemmateam2025gemma3technicalreport}. For specialized data classification tasks within this framework, we employed Small Language Models (SLMs), leveraging their established effectiveness and low latency, which helps to quickly run policies in real-time. Both policies were judged \texttt{request review} by security engineers as part of our evaluation and deployed after minimal changes, saving time for the security team. \texttt{POLICY-AS-PROMPT-GEN} was evaluated on functional correctness by integrating generated prompts into target LLMs and testing with 100 gold inputs and outputs. Tests included both standard and adversarial cases (e.g., prompt injection, toxic content). Evaluation focused on accuracy of \texttt{OOD}/\texttt{ID} classifications, with defaults of \texttt{OOD} $\Rightarrow$ \textbf{BLOCK} and \texttt{ID} $\Rightarrow$ \textbf{ALLOW}, unless otherwise specified. While accuracies in the ~70-73\% range for \texttt{GPT-4o} (Table~\ref{tab:acc-app-model}) are not perfect, they demonstrate significant utility in a real-world context. The system acts as a "default-deny" guardrail: its primary value is in successfully blocking a high percentage of malicious or non-compliant inputs (true positives for \texttt{OOD}) before they reach the agent, drastically reducing the attack surface. This level of accuracy is highly effective as a first-line defense, flagging ambiguous cases for human review, which is a significant improvement over manual, post-hoc auditing. A limitation is that prompt tuning emphasized GPT models, potentially contributing to performance gaps.

\section{Conclusion}
In this work, we introduced a framework that bridges the policy-to-practice gap by transforming unstructured design artifacts into verifiable guardrails. We demonstrated an end-to-end pipeline for automated, auditable, and enforceable policy generation.  
Our experiments show that while large proprietary models excel in policy extraction, smaller models can still effectively enforce policies with curated prompts.  
This approach establishes a scalable path toward trustworthy, regulatable AI systems grounded in transparent policy governance.

\section{Limitations}
\label{sec:limitations}
Despite promising results, this work has several limitations that temper generalizability and reproducibility. Our evaluation spans two enterprise-style domains (HR, SOC) with modest gold sets (100 inputs and 100 outputs per application) and a limited number of design artifacts, which constrains transferability to other settings and larger, more heterogeneous corpora. Policy extraction currently depends on large proprietary models, while open-source baselines differ in size or tuning; moreover, prompting effort was greater for GPT-family models, potentially biasing results and limiting apples-to-apples comparisons. Reproducibility is further hindered by confidentiality of internal artifacts and logs: we cannot release the full corpora. Future work includes leveraging production interaction logs (inputs, outputs, tool calls, actions) to mine candidate rules and hard examples that continuously enrich the policy tree and prompts, and implementing adaptive policy regeneration that automatically re-runs \texttt{POLICY-TREE-GEN} and \texttt{POLICY-AS-PROMPT-GEN} whenever PRDs, TDDs, code, or schemas undergo significant changes, with versioning and regression gating to mitigate drift. Also, prompt optimisations like \cite{kumar2025sculptsystematictuninglong} can be applied to \texttt{POLICY-AS-PROMPT}.

\bibliographystyle{plainnat}   
\bibliography{refs}            

\begin{thebibliography}{20}
\providecommand{\natexlab}[1]{#1}
\providecommand{\url}[1]{\texttt{#1}}
\expandafter\ifx\csname urlstyle\endcsname\relax
  \providecommand{\doi}[1]{doi: #1}\else
  \providecommand{\doi}{doi: \begingroup \urlstyle{rm}\Url}\fi

\bibitem[eu-(2024)]{eu-ai-act}
Regulation (eu) 2024/1689 of the european parliament and of the council laying down harmonised rules on artificial intelligence (artificial intelligence act).
\newblock Official Journal of the European Union, L 298, 12 July 2024, 2024.
\newblock URL \url{https://artificialintelligenceact.eu/ai-act-explorer/}.
\newblock Accessible via the AI Act Explorer.

\bibitem[{Anthropic}(2024)]{anthropic_claude35_sonnet_2024}
{Anthropic}.
\newblock Introducing claude 3.5 sonnet.
\newblock \url{https://www.anthropic.com/news/claude-3-5-sonnet}, June 2024.
\newblock Blog post, published June 20 2024; accessed [insert access date].

\bibitem[Ding et~al.(2024)Ding, Li, Wang, and Chen]{ding2024large}
Han Ding, Yinheng Li, Junhao Wang, and Hang Chen.
\newblock Large language model agent in financial trading: A survey.
\newblock \emph{arXiv preprint arXiv:2408.06361}, 2024.

\bibitem[Dong et~al.(2024)Dong, Mu, Jin, Qi, Hu, Zhao, Meng, Ruan, and Huang]{dong2024building}
Yi~Dong, Ronghui Mu, Gaojie Jin, Yi~Qi, Jinwei Hu, Xingyu Zhao, Jie Meng, Wenjie Ruan, and Xiaowei Huang.
\newblock Building guardrails for large language models.
\newblock \emph{arXiv preprint arXiv:2402.01822}, 2024.

\bibitem[Fabiano(2024)]{fabiano2024aiactlargelanguage}
Nicola Fabiano.
\newblock Ai act and large language models (llms): When critical issues and privacy impact require human and ethical oversight, 2024.
\newblock URL \url{https://arxiv.org/abs/2404.00600}.

\bibitem[Grattafiori et~al.(2024)Grattafiori, Dubey, Jauhri, Pandey, Kadian, Al-Dahle, Letman, Mathur, Schelten, Vaughan, Yang, Fan, Goyal, Hartshorn, Yang, Mitra, Sravankumar, Korenev, Hinsvark, Rao, Zhang, Rodriguez, Gregerson, Spataru, Roziere, Biron, Tang, Chern, Caucheteux, Nayak, Bi, Marra, McConnell, Keller, Touret, Wu, Wong, Ferrer, Nikolaidis, Allonsius, Song, Pintz, Livshits, Wyatt, Esiobu, Choudhary, Mahajan, Garcia-Olano, Perino, Hupkes, Lakomkin, AlBadawy, Lobanova, Dinan, Smith, Radenovic, Guzmán, Zhang, Synnaeve, Lee, Anderson, Thattai, Nail, Mialon, Pang, Cucurell, Nguyen, Korevaar, Xu, Touvron, Zarov, Ibarra, Kloumann, Misra, Evtimov, Zhang, Copet, Lee, Geffert, Vranes, Park, Mahadeokar, Shah, van~der Linde, Billock, Hong, Lee, Fu, Chi, Huang, Liu, Wang, Yu, Bitton, Spisak, Park, Rocca, Johnstun, Saxe, Jia, Alwala, Prasad, Upasani, Plawiak, Li, Heafield, Stone, El-Arini, Iyer, Malik, Chiu, Bhalla, Lakhotia, Rantala-Yeary, van~der Maaten, Chen, Tan, Jenkins, Martin, Madaan, Malo, Blecher,
  Landzaat, de~Oliveira, Muzzi, Pasupuleti, Singh, Paluri, Kardas, Tsimpoukelli, Oldham, Rita, Pavlova, Kambadur, Lewis, Si, Singh, Hassan, Goyal, Torabi, Bashlykov, Bogoychev, Chatterji, Zhang, Duchenne, Çelebi, Alrassy, Zhang, Li, Vasic, Weng, Bhargava, Dubal, Krishnan, Koura, Xu, He, Dong, Srinivasan, Ganapathy, Calderer, Cabral, Stojnic, Raileanu, Maheswari, Girdhar, Patel, Sauvestre, Polidoro, Sumbaly, Taylor, Silva, Hou, Wang, Hosseini, Chennabasappa, Singh, Bell, Kim, Edunov, Nie, Narang, Raparthy, Shen, Wan, Bhosale, Zhang, Vandenhende, Batra, Whitman, Sootla, Collot, Gururangan, Borodinsky, Herman, Fowler, Sheasha, Georgiou, Scialom, Speckbacher, Mihaylov, Xiao, Karn, Goswami, Gupta, Ramanathan, Kerkez, Gonguet, Do, Vogeti, Albiero, Petrovic, Chu, Xiong, Fu, Meers, Martinet, Wang, Wang, Tan, Xia, Xie, Jia, Wang, Goldschlag, Gaur, Babaei, Wen, Song, Zhang, Li, Mao, Coudert, Yan, Chen, Papakipos, Singh, Srivastava, Jain, Kelsey, Shajnfeld, Gangidi, Victoria, Goldstand, Menon, Sharma, Boesenberg,
  Baevski, Feinstein, Kallet, Sangani, Teo, Yunus, Lupu, Alvarado, Caples, Gu, Ho, Poulton, Ryan, Ramchandani, Dong, Franco, Goyal, Saraf, Chowdhury, Gabriel, Bharambe, Eisenman, Yazdan, James, Maurer, Leonhardi, Huang, Loyd, Paola, Paranjape, Liu, Wu, Ni, Hancock, Wasti, Spence, Stojkovic, Gamido, Montalvo, Parker, Burton, Mejia, Liu, Wang, Kim, Zhou, Hu, Chu, Cai, Tindal, Feichtenhofer, Gao, Civin, Beaty, Kreymer, Li, Adkins, Xu, Testuggine, David, Parikh, Liskovich, Foss, Wang, Le, Holland, Dowling, Jamil, Montgomery, Presani, Hahn, Wood, Le, Brinkman, Arcaute, Dunbar, Smothers, Sun, Kreuk, Tian, Kokkinos, Ozgenel, Caggioni, Kanayet, Seide, Florez, Schwarz, Badeer, Swee, Halpern, Herman, Sizov, Guangyi, Zhang, Lakshminarayanan, Inan, Shojanazeri, Zou, Wang, Zha, Habeeb, Rudolph, Suk, Aspegren, Goldman, Zhan, Damlaj, Molybog, Tufanov, Leontiadis, Veliche, Gat, Weissman, Geboski, Kohli, Lam, Asher, Gaya, Marcus, Tang, Chan, Zhen, Reizenstein, Teboul, Zhong, Jin, Yang, Cummings, Carvill, Shepard, McPhie,
  Torres, Ginsburg, Wang, Wu, U, Saxena, Khandelwal, Zand, Matosich, Veeraraghavan, Michelena, Li, Jagadeesh, Huang, Chawla, Huang, Chen, Garg, A, Silva, Bell, Zhang, Guo, Yu, Moshkovich, Wehrstedt, Khabsa, Avalani, Bhatt, Mankus, Hasson, Lennie, Reso, Groshev, Naumov, Lathi, Keneally, Liu, Seltzer, Valko, Restrepo, Patel, Vyatskov, Samvelyan, Clark, Macey, Wang, Hermoso, Metanat, Rastegari, Bansal, Santhanam, Parks, White, Bawa, Singhal, Egebo, Usunier, Mehta, Laptev, Dong, Cheng, Chernoguz, Hart, Salpekar, Kalinli, Kent, Parekh, Saab, Balaji, Rittner, Bontrager, Roux, Dollar, Zvyagina, Ratanchandani, Yuvraj, Liang, Alao, Rodriguez, Ayub, Murthy, Nayani, Mitra, Parthasarathy, Li, Hogan, Battey, Wang, Howes, Rinott, Mehta, Siby, Bondu, Datta, Chugh, Hunt, Dhillon, Sidorov, Pan, Mahajan, Verma, Yamamoto, Ramaswamy, Lindsay, Lindsay, Feng, Lin, Zha, Patil, Shankar, Zhang, Zhang, Wang, Agarwal, Sajuyigbe, Chintala, Max, Chen, Kehoe, Satterfield, Govindaprasad, Gupta, Deng, Cho, Virk, Subramanian, Choudhury,
  Goldman, Remez, Glaser, Best, Koehler, Robinson, Li, Zhang, Matthews, Chou, Shaked, Vontimitta, Ajayi, Montanez, Mohan, Kumar, Mangla, Ionescu, Poenaru, Mihailescu, Ivanov, Li, Wang, Jiang, Bouaziz, Constable, Tang, Wu, Wang, Wu, Gao, Kleinman, Chen, Hu, Jia, Qi, Li, Zhang, Zhang, Adi, Nam, Yu, Wang, Zhao, Hao, Qian, Li, He, Rait, DeVito, Rosnbrick, Wen, Yang, Zhao, and Ma]{grattafiori2024llama3herdmodels}
Aaron Grattafiori, Abhimanyu Dubey, Abhinav Jauhri, Abhinav Pandey, Abhishek Kadian, Ahmad Al-Dahle, Aiesha Letman, Akhil Mathur, Alan Schelten, Alex Vaughan, Amy Yang, Angela Fan, Anirudh Goyal, Anthony Hartshorn, Aobo Yang, Archi Mitra, Archie Sravankumar, Artem Korenev, Arthur Hinsvark, Arun Rao, Aston Zhang, Aurelien Rodriguez, Austen Gregerson, Ava Spataru, Baptiste Roziere, Bethany Biron, Binh Tang, Bobbie Chern, Charlotte Caucheteux, Chaya Nayak, Chloe Bi, Chris Marra, Chris McConnell, Christian Keller, Christophe Touret, Chunyang Wu, Corinne Wong, Cristian~Canton Ferrer, Cyrus Nikolaidis, Damien Allonsius, Daniel Song, Danielle Pintz, Danny Livshits, Danny Wyatt, David Esiobu, Dhruv Choudhary, Dhruv Mahajan, Diego Garcia-Olano, Diego Perino, Dieuwke Hupkes, Egor Lakomkin, Ehab AlBadawy, Elina Lobanova, Emily Dinan, Eric~Michael Smith, Filip Radenovic, Francisco Guzmán, Frank Zhang, Gabriel Synnaeve, Gabrielle Lee, Georgia~Lewis Anderson, Govind Thattai, Graeme Nail, Gregoire Mialon, Guan Pang,
  Guillem Cucurell, Hailey Nguyen, Hannah Korevaar, Hu~Xu, Hugo Touvron, Iliyan Zarov, Imanol~Arrieta Ibarra, Isabel Kloumann, Ishan Misra, Ivan Evtimov, Jack Zhang, Jade Copet, Jaewon Lee, Jan Geffert, Jana Vranes, Jason Park, Jay Mahadeokar, Jeet Shah, Jelmer van~der Linde, Jennifer Billock, Jenny Hong, Jenya Lee, Jeremy Fu, Jianfeng Chi, Jianyu Huang, Jiawen Liu, Jie Wang, Jiecao Yu, Joanna Bitton, Joe Spisak, Jongsoo Park, Joseph Rocca, Joshua Johnstun, Joshua Saxe, Junteng Jia, Kalyan~Vasuden Alwala, Karthik Prasad, Kartikeya Upasani, Kate Plawiak, Ke~Li, Kenneth Heafield, Kevin Stone, Khalid El-Arini, Krithika Iyer, Kshitiz Malik, Kuenley Chiu, Kunal Bhalla, Kushal Lakhotia, Lauren Rantala-Yeary, Laurens van~der Maaten, Lawrence Chen, Liang Tan, Liz Jenkins, Louis Martin, Lovish Madaan, Lubo Malo, Lukas Blecher, Lukas Landzaat, Luke de~Oliveira, Madeline Muzzi, Mahesh Pasupuleti, Mannat Singh, Manohar Paluri, Marcin Kardas, Maria Tsimpoukelli, Mathew Oldham, Mathieu Rita, Maya Pavlova, Melanie Kambadur,
  Mike Lewis, Min Si, Mitesh~Kumar Singh, Mona Hassan, Naman Goyal, Narjes Torabi, Nikolay Bashlykov, Nikolay Bogoychev, Niladri Chatterji, Ning Zhang, Olivier Duchenne, Onur Çelebi, Patrick Alrassy, Pengchuan Zhang, Pengwei Li, Petar Vasic, Peter Weng, Prajjwal Bhargava, Pratik Dubal, Praveen Krishnan, Punit~Singh Koura, Puxin Xu, Qing He, Qingxiao Dong, Ragavan Srinivasan, Raj Ganapathy, Ramon Calderer, Ricardo~Silveira Cabral, Robert Stojnic, Roberta Raileanu, Rohan Maheswari, Rohit Girdhar, Rohit Patel, Romain Sauvestre, Ronnie Polidoro, Roshan Sumbaly, Ross Taylor, Ruan Silva, Rui Hou, Rui Wang, Saghar Hosseini, Sahana Chennabasappa, Sanjay Singh, Sean Bell, Seohyun~Sonia Kim, Sergey Edunov, Shaoliang Nie, Sharan Narang, Sharath Raparthy, Sheng Shen, Shengye Wan, Shruti Bhosale, Shun Zhang, Simon Vandenhende, Soumya Batra, Spencer Whitman, Sten Sootla, Stephane Collot, Suchin Gururangan, Sydney Borodinsky, Tamar Herman, Tara Fowler, Tarek Sheasha, Thomas Georgiou, Thomas Scialom, Tobias Speckbacher,
  Todor Mihaylov, Tong Xiao, Ujjwal Karn, Vedanuj Goswami, Vibhor Gupta, Vignesh Ramanathan, Viktor Kerkez, Vincent Gonguet, Virginie Do, Vish Vogeti, Vítor Albiero, Vladan Petrovic, Weiwei Chu, Wenhan Xiong, Wenyin Fu, Whitney Meers, Xavier Martinet, Xiaodong Wang, Xiaofang Wang, Xiaoqing~Ellen Tan, Xide Xia, Xinfeng Xie, Xuchao Jia, Xuewei Wang, Yaelle Goldschlag, Yashesh Gaur, Yasmine Babaei, Yi~Wen, Yiwen Song, Yuchen Zhang, Yue Li, Yuning Mao, Zacharie~Delpierre Coudert, Zheng Yan, Zhengxing Chen, Zoe Papakipos, Aaditya Singh, Aayushi Srivastava, Abha Jain, Adam Kelsey, Adam Shajnfeld, Adithya Gangidi, Adolfo Victoria, Ahuva Goldstand, Ajay Menon, Ajay Sharma, Alex Boesenberg, Alexei Baevski, Allie Feinstein, Amanda Kallet, Amit Sangani, Amos Teo, Anam Yunus, Andrei Lupu, Andres Alvarado, Andrew Caples, Andrew Gu, Andrew Ho, Andrew Poulton, Andrew Ryan, Ankit Ramchandani, Annie Dong, Annie Franco, Anuj Goyal, Aparajita Saraf, Arkabandhu Chowdhury, Ashley Gabriel, Ashwin Bharambe, Assaf Eisenman, Azadeh
  Yazdan, Beau James, Ben Maurer, Benjamin Leonhardi, Bernie Huang, Beth Loyd, Beto~De Paola, Bhargavi Paranjape, Bing Liu, Bo~Wu, Boyu Ni, Braden Hancock, Bram Wasti, Brandon Spence, Brani Stojkovic, Brian Gamido, Britt Montalvo, Carl Parker, Carly Burton, Catalina Mejia, Ce~Liu, Changhan Wang, Changkyu Kim, Chao Zhou, Chester Hu, Ching-Hsiang Chu, Chris Cai, Chris Tindal, Christoph Feichtenhofer, Cynthia Gao, Damon Civin, Dana Beaty, Daniel Kreymer, Daniel Li, David Adkins, David Xu, Davide Testuggine, Delia David, Devi Parikh, Diana Liskovich, Didem Foss, Dingkang Wang, Duc Le, Dustin Holland, Edward Dowling, Eissa Jamil, Elaine Montgomery, Eleonora Presani, Emily Hahn, Emily Wood, Eric-Tuan Le, Erik Brinkman, Esteban Arcaute, Evan Dunbar, Evan Smothers, Fei Sun, Felix Kreuk, Feng Tian, Filippos Kokkinos, Firat Ozgenel, Francesco Caggioni, Frank Kanayet, Frank Seide, Gabriela~Medina Florez, Gabriella Schwarz, Gada Badeer, Georgia Swee, Gil Halpern, Grant Herman, Grigory Sizov, Guangyi, Zhang, Guna
  Lakshminarayanan, Hakan Inan, Hamid Shojanazeri, Han Zou, Hannah Wang, Hanwen Zha, Haroun Habeeb, Harrison Rudolph, Helen Suk, Henry Aspegren, Hunter Goldman, Hongyuan Zhan, Ibrahim Damlaj, Igor Molybog, Igor Tufanov, Ilias Leontiadis, Irina-Elena Veliche, Itai Gat, Jake Weissman, James Geboski, James Kohli, Janice Lam, Japhet Asher, Jean-Baptiste Gaya, Jeff Marcus, Jeff Tang, Jennifer Chan, Jenny Zhen, Jeremy Reizenstein, Jeremy Teboul, Jessica Zhong, Jian Jin, Jingyi Yang, Joe Cummings, Jon Carvill, Jon Shepard, Jonathan McPhie, Jonathan Torres, Josh Ginsburg, Junjie Wang, Kai Wu, Kam~Hou U, Karan Saxena, Kartikay Khandelwal, Katayoun Zand, Kathy Matosich, Kaushik Veeraraghavan, Kelly Michelena, Keqian Li, Kiran Jagadeesh, Kun Huang, Kunal Chawla, Kyle Huang, Lailin Chen, Lakshya Garg, Lavender A, Leandro Silva, Lee Bell, Lei Zhang, Liangpeng Guo, Licheng Yu, Liron Moshkovich, Luca Wehrstedt, Madian Khabsa, Manav Avalani, Manish Bhatt, Martynas Mankus, Matan Hasson, Matthew Lennie, Matthias Reso, Maxim
  Groshev, Maxim Naumov, Maya Lathi, Meghan Keneally, Miao Liu, Michael~L. Seltzer, Michal Valko, Michelle Restrepo, Mihir Patel, Mik Vyatskov, Mikayel Samvelyan, Mike Clark, Mike Macey, Mike Wang, Miquel~Jubert Hermoso, Mo~Metanat, Mohammad Rastegari, Munish Bansal, Nandhini Santhanam, Natascha Parks, Natasha White, Navyata Bawa, Nayan Singhal, Nick Egebo, Nicolas Usunier, Nikhil Mehta, Nikolay~Pavlovich Laptev, Ning Dong, Norman Cheng, Oleg Chernoguz, Olivia Hart, Omkar Salpekar, Ozlem Kalinli, Parkin Kent, Parth Parekh, Paul Saab, Pavan Balaji, Pedro Rittner, Philip Bontrager, Pierre Roux, Piotr Dollar, Polina Zvyagina, Prashant Ratanchandani, Pritish Yuvraj, Qian Liang, Rachad Alao, Rachel Rodriguez, Rafi Ayub, Raghotham Murthy, Raghu Nayani, Rahul Mitra, Rangaprabhu Parthasarathy, Raymond Li, Rebekkah Hogan, Robin Battey, Rocky Wang, Russ Howes, Ruty Rinott, Sachin Mehta, Sachin Siby, Sai~Jayesh Bondu, Samyak Datta, Sara Chugh, Sara Hunt, Sargun Dhillon, Sasha Sidorov, Satadru Pan, Saurabh Mahajan,
  Saurabh Verma, Seiji Yamamoto, Sharadh Ramaswamy, Shaun Lindsay, Shaun Lindsay, Sheng Feng, Shenghao Lin, Shengxin~Cindy Zha, Shishir Patil, Shiva Shankar, Shuqiang Zhang, Shuqiang Zhang, Sinong Wang, Sneha Agarwal, Soji Sajuyigbe, Soumith Chintala, Stephanie Max, Stephen Chen, Steve Kehoe, Steve Satterfield, Sudarshan Govindaprasad, Sumit Gupta, Summer Deng, Sungmin Cho, Sunny Virk, Suraj Subramanian, Sy~Choudhury, Sydney Goldman, Tal Remez, Tamar Glaser, Tamara Best, Thilo Koehler, Thomas Robinson, Tianhe Li, Tianjun Zhang, Tim Matthews, Timothy Chou, Tzook Shaked, Varun Vontimitta, Victoria Ajayi, Victoria Montanez, Vijai Mohan, Vinay~Satish Kumar, Vishal Mangla, Vlad Ionescu, Vlad Poenaru, Vlad~Tiberiu Mihailescu, Vladimir Ivanov, Wei Li, Wenchen Wang, Wenwen Jiang, Wes Bouaziz, Will Constable, Xiaocheng Tang, Xiaojian Wu, Xiaolan Wang, Xilun Wu, Xinbo Gao, Yaniv Kleinman, Yanjun Chen, Ye~Hu, Ye~Jia, Ye~Qi, Yenda Li, Yilin Zhang, Ying Zhang, Yossi Adi, Youngjin Nam, Yu, Wang, Yu~Zhao, Yuchen Hao, Yundi
  Qian, Yunlu Li, Yuzi He, Zach Rait, Zachary DeVito, Zef Rosnbrick, Zhaoduo Wen, Zhenyu Yang, Zhiwei Zhao, and Zhiyu Ma.
\newblock The llama 3 herd of models, 2024.
\newblock URL \url{https://arxiv.org/abs/2407.21783}.

\bibitem[He et~al.(2024)He, Wang, Rong, Cheng, and Chen]{he2024securityaiagents}
Yifeng He, Ethan Wang, Yuyang Rong, Zifei Cheng, and Hao Chen.
\newblock Security of ai agents, 2024.
\newblock URL \url{https://arxiv.org/abs/2406.08689}.

\bibitem[Hua et~al.(2024)Hua, Yang, Jin, Li, Cheng, Tang, and Zhang]{hua2024trustagentsafetrustworthyllmbased}
Wenyue Hua, Xianjun Yang, Mingyu Jin, Zelong Li, Wei Cheng, Ruixiang Tang, and Yongfeng Zhang.
\newblock Trustagent: Towards safe and trustworthy llm-based agents, 2024.
\newblock URL \url{https://arxiv.org/abs/2402.01586}.

\bibitem[Kholkar and Ahuja(2025)]{kholkar2025capturecontextawarepromptinjection}
Gauri Kholkar and Ratinder Ahuja.
\newblock Capture: Context-aware prompt injection testing and robustness enhancement, 2025.
\newblock URL \url{https://arxiv.org/abs/2505.12368}.

\bibitem[Kim et~al.(2023)Kim, Yun, Lee, Gubri, Yoon, and Oh]{kim2023propileprobingprivacyleakage}
Siwon Kim, Sangdoo Yun, Hwaran Lee, Martin Gubri, Sungroh Yoon, and Seong~Joon Oh.
\newblock Propile: Probing privacy leakage in large language models, 2023.
\newblock URL \url{https://arxiv.org/abs/2307.01881}.

\bibitem[Kumar et~al.(2025)Kumar, Venkata, Khandelwal, Santra, Agrawal, and Gupta]{kumar2025sculptsystematictuninglong}
Shanu Kumar, Akhila~Yesantarao Venkata, Shubhanshu Khandelwal, Bishal Santra, Parag Agrawal, and Manish Gupta.
\newblock Sculpt: Systematic tuning of long prompts, 2025.
\newblock URL \url{https://arxiv.org/abs/2410.20788}.

\bibitem[Li et~al.(2025)Li, Zhou, Raghuram, Goldstein, and Goldblum]{li2025commercialllmagentsvulnerable}
Ang Li, Yin Zhou, Vethavikashini~Chithrra Raghuram, Tom Goldstein, and Micah Goldblum.
\newblock Commercial llm agents are already vulnerable to simple yet dangerous attacks, 2025.
\newblock URL \url{https://arxiv.org/abs/2502.08586}.

\bibitem[Liu et~al.(2024)Liu, Deng, Li, Wang, Wang, Wang, Zhang, Liu, Wang, Zheng, and Liu]{liu2024promptinjectionattackllmintegrated}
Yi~Liu, Gelei Deng, Yuekang Li, Kailong Wang, Zihao Wang, Xiaofeng Wang, Tianwei Zhang, Yepang Liu, Haoyu Wang, Yan Zheng, and Yang Liu.
\newblock Prompt injection attack against llm-integrated applications, 2024.
\newblock URL \url{https://arxiv.org/abs/2306.05499}.

\bibitem[OpenAI et~al.(2024)OpenAI, :, Jaech, Kalai, Lerer, Richardson, El-Kishky, Low, Helyar, Madry, Beutel, Carney, Iftimie, Karpenko, Passos, Neitz, Prokofiev, Wei, Tam, Bennett, Kumar, Saraiva, Vallone, Duberstein, Kondrich, Mishchenko, Applebaum, Jiang, Nair, Zoph, Ghorbani, Rossen, Sokolowsky, Barak, McGrew, Minaiev, Hao, Baker, Houghton, McKinzie, Eastman, Lugaresi, Bassin, Hudson, Li, de~Bourcy, Voss, Shen, Zhang, Koch, Orsinger, Hesse, Fischer, Chan, Roberts, Kappler, Levy, Selsam, Dohan, Farhi, Mely, Robinson, Tsipras, Li, Oprica, Freeman, Zhang, Wong, Proehl, Cheung, Mitchell, Wallace, Ritter, Mays, Wang, Such, Raso, Leoni, Tsimpourlas, Song, von Lohmann, Sulit, Salmon, Parascandolo, Chabot, Zhao, Brockman, Leclerc, Salman, Bao, Sheng, Andrin, Bagherinezhad, Ren, Lightman, Chung, Kivlichan, O'Connell, Osband, Gilaberte, Akkaya, Kostrikov, Sutskever, Kofman, Pachocki, Lennon, Wei, Harb, Twore, Feng, Yu, Weng, Tang, Yu, Candela, Palermo, Parish, Heidecke, Hallman, Rizzo, Gordon, Uesato, Ward,
  Huizinga, Wang, Chen, Xiao, Singhal, Nguyen, Cobbe, Shi, Wood, Rimbach, Gu-Lemberg, Liu, Lu, Stone, Yu, Ahmad, Yang, Liu, Maksin, Ho, Fedus, Weng, Li, McCallum, Held, Kuhn, Kondraciuk, Kaiser, Metz, Boyd, Trebacz, Joglekar, Chen, Tintor, Meyer, Jones, Kaufer, Schwarzer, Shah, Yatbaz, Guan, Xu, Yan, Glaese, Chen, Lampe, Malek, Wang, Fradin, McClay, Pavlov, Wang, Wang, Murati, Bavarian, Rohaninejad, McAleese, Chowdhury, Chowdhury, Ryder, Tezak, Brown, Nachum, Boiko, Murk, Watkins, Chao, Ashbourne, Izmailov, Zhokhov, Dias, Arora, Lin, Lopes, Gaon, Miyara, Leike, Hwang, Garg, Brown, James, Shu, Cheu, Greene, Jain, Altman, Toizer, Toyer, Miserendino, Agarwal, Hernandez, Baker, McKinney, Yan, Zhao, Hu, Santurkar, Chaudhuri, Zhang, Fu, Papay, Lin, Balaji, Sanjeev, Sidor, Broda, Clark, Wang, Gordon, Sanders, Patwardhan, Sottiaux, Degry, Dimson, Zheng, Garipov, Stasi, Bansal, Creech, Peterson, Eloundou, Qi, Kosaraju, Monaco, Pong, Fomenko, Zheng, Zhou, McCabe, Zaremba, Dubois, Lu, Chen, Cha, Bai, He, Zhang, Wang,
  Shao, and Li]{openai2024openaio1card}
OpenAI, :, Aaron Jaech, Adam Kalai, Adam Lerer, Adam Richardson, Ahmed El-Kishky, Aiden Low, Alec Helyar, Aleksander Madry, Alex Beutel, Alex Carney, Alex Iftimie, Alex Karpenko, Alex~Tachard Passos, Alexander Neitz, Alexander Prokofiev, Alexander Wei, Allison Tam, Ally Bennett, Ananya Kumar, Andre Saraiva, Andrea Vallone, Andrew Duberstein, Andrew Kondrich, Andrey Mishchenko, Andy Applebaum, Angela Jiang, Ashvin Nair, Barret Zoph, Behrooz Ghorbani, Ben Rossen, Benjamin Sokolowsky, Boaz Barak, Bob McGrew, Borys Minaiev, Botao Hao, Bowen Baker, Brandon Houghton, Brandon McKinzie, Brydon Eastman, Camillo Lugaresi, Cary Bassin, Cary Hudson, Chak~Ming Li, Charles de~Bourcy, Chelsea Voss, Chen Shen, Chong Zhang, Chris Koch, Chris Orsinger, Christopher Hesse, Claudia Fischer, Clive Chan, Dan Roberts, Daniel Kappler, Daniel Levy, Daniel Selsam, David Dohan, David Farhi, David Mely, David Robinson, Dimitris Tsipras, Doug Li, Dragos Oprica, Eben Freeman, Eddie Zhang, Edmund Wong, Elizabeth Proehl, Enoch Cheung, Eric
  Mitchell, Eric Wallace, Erik Ritter, Evan Mays, Fan Wang, Felipe~Petroski Such, Filippo Raso, Florencia Leoni, Foivos Tsimpourlas, Francis Song, Fred von Lohmann, Freddie Sulit, Geoff Salmon, Giambattista Parascandolo, Gildas Chabot, Grace Zhao, Greg Brockman, Guillaume Leclerc, Hadi Salman, Haiming Bao, Hao Sheng, Hart Andrin, Hessam Bagherinezhad, Hongyu Ren, Hunter Lightman, Hyung~Won Chung, Ian Kivlichan, Ian O'Connell, Ian Osband, Ignasi~Clavera Gilaberte, Ilge Akkaya, Ilya Kostrikov, Ilya Sutskever, Irina Kofman, Jakub Pachocki, James Lennon, Jason Wei, Jean Harb, Jerry Twore, Jiacheng Feng, Jiahui Yu, Jiayi Weng, Jie Tang, Jieqi Yu, Joaquin~Quiñonero Candela, Joe Palermo, Joel Parish, Johannes Heidecke, John Hallman, John Rizzo, Jonathan Gordon, Jonathan Uesato, Jonathan Ward, Joost Huizinga, Julie Wang, Kai Chen, Kai Xiao, Karan Singhal, Karina Nguyen, Karl Cobbe, Katy Shi, Kayla Wood, Kendra Rimbach, Keren Gu-Lemberg, Kevin Liu, Kevin Lu, Kevin Stone, Kevin Yu, Lama Ahmad, Lauren Yang, Leo Liu,
  Leon Maksin, Leyton Ho, Liam Fedus, Lilian Weng, Linden Li, Lindsay McCallum, Lindsey Held, Lorenz Kuhn, Lukas Kondraciuk, Lukasz Kaiser, Luke Metz, Madelaine Boyd, Maja Trebacz, Manas Joglekar, Mark Chen, Marko Tintor, Mason Meyer, Matt Jones, Matt Kaufer, Max Schwarzer, Meghan Shah, Mehmet Yatbaz, Melody~Y. Guan, Mengyuan Xu, Mengyuan Yan, Mia Glaese, Mianna Chen, Michael Lampe, Michael Malek, Michele Wang, Michelle Fradin, Mike McClay, Mikhail Pavlov, Miles Wang, Mingxuan Wang, Mira Murati, Mo~Bavarian, Mostafa Rohaninejad, Nat McAleese, Neil Chowdhury, Neil Chowdhury, Nick Ryder, Nikolas Tezak, Noam Brown, Ofir Nachum, Oleg Boiko, Oleg Murk, Olivia Watkins, Patrick Chao, Paul Ashbourne, Pavel Izmailov, Peter Zhokhov, Rachel Dias, Rahul Arora, Randall Lin, Rapha~Gontijo Lopes, Raz Gaon, Reah Miyara, Reimar Leike, Renny Hwang, Rhythm Garg, Robin Brown, Roshan James, Rui Shu, Ryan Cheu, Ryan Greene, Saachi Jain, Sam Altman, Sam Toizer, Sam Toyer, Samuel Miserendino, Sandhini Agarwal, Santiago Hernandez,
  Sasha Baker, Scott McKinney, Scottie Yan, Shengjia Zhao, Shengli Hu, Shibani Santurkar, Shraman~Ray Chaudhuri, Shuyuan Zhang, Siyuan Fu, Spencer Papay, Steph Lin, Suchir Balaji, Suvansh Sanjeev, Szymon Sidor, Tal Broda, Aidan Clark, Tao Wang, Taylor Gordon, Ted Sanders, Tejal Patwardhan, Thibault Sottiaux, Thomas Degry, Thomas Dimson, Tianhao Zheng, Timur Garipov, Tom Stasi, Trapit Bansal, Trevor Creech, Troy Peterson, Tyna Eloundou, Valerie Qi, Vineet Kosaraju, Vinnie Monaco, Vitchyr Pong, Vlad Fomenko, Weiyi Zheng, Wenda Zhou, Wes McCabe, Wojciech Zaremba, Yann Dubois, Yinghai Lu, Yining Chen, Young Cha, Yu~Bai, Yuchen He, Yuchen Zhang, Yunyun Wang, Zheng Shao, and Zhuohan Li.
\newblock Openai o1 system card, 2024.
\newblock URL \url{https://arxiv.org/abs/2412.16720}.

\bibitem[OpenAI et~al.(2025)OpenAI, :, Agarwal, Ahmad, Ai, Altman, Applebaum, Arbus, Arora, Bai, Baker, Bao, Barak, Bennett, Bertao, Brett, Brevdo, Brockman, Bubeck, Chang, Chen, Chen, Cheung, Clark, Cook, Dukhan, Dvorak, Fives, Fomenko, Garipov, Georgiev, Glaese, Gogineni, Goucher, Gross, Guzman, Hallman, Hehir, Heidecke, Helyar, Hu, Huet, Huh, Jain, Johnson, Koch, Kofman, Kundel, Kwon, Kyrylov, Le, Leclerc, Lennon, Lessans, Lezcano-Casado, Li, Li, Lin, Liss, Lily, Liu, Liu, Lu, Lu, Martinovic, McCallum, McGrath, McKinney, McLaughlin, Mei, Mostovoy, Mu, Myles, Neitz, Nichol, Pachocki, Paino, Palmie, Pantuliano, Parascandolo, Park, Pathak, Paz, Peran, Pimenov, Pokrass, Proehl, Qiu, Raila, Raso, Ren, Richardson, Robinson, Rotsted, Salman, Sanjeev, Schwarzer, Sculley, Sikchi, Simon, Singhal, Song, Stuckey, Sun, Tillet, Toizer, Tsimpourlas, Vyas, Wallace, Wang, Wang, Watkins, Weil, Wendling, Whinnery, Whitney, Wong, Yang, Yang, Yasunaga, Ying, Zaremba, Zhan, Zhang, Zhang, Zhang, and
  Zhao]{openai2025gptoss120bgptoss20bmodel}
OpenAI, :, Sandhini Agarwal, Lama Ahmad, Jason Ai, Sam Altman, Andy Applebaum, Edwin Arbus, Rahul~K. Arora, Yu~Bai, Bowen Baker, Haiming Bao, Boaz Barak, Ally Bennett, Tyler Bertao, Nivedita Brett, Eugene Brevdo, Greg Brockman, Sebastien Bubeck, Che Chang, Kai Chen, Mark Chen, Enoch Cheung, Aidan Clark, Dan Cook, Marat Dukhan, Casey Dvorak, Kevin Fives, Vlad Fomenko, Timur Garipov, Kristian Georgiev, Mia Glaese, Tarun Gogineni, Adam Goucher, Lukas Gross, Katia~Gil Guzman, John Hallman, Jackie Hehir, Johannes Heidecke, Alec Helyar, Haitang Hu, Romain Huet, Jacob Huh, Saachi Jain, Zach Johnson, Chris Koch, Irina Kofman, Dominik Kundel, Jason Kwon, Volodymyr Kyrylov, Elaine~Ya Le, Guillaume Leclerc, James~Park Lennon, Scott Lessans, Mario Lezcano-Casado, Yuanzhi Li, Zhuohan Li, Ji~Lin, Jordan Liss, Lily, Liu, Jiancheng Liu, Kevin Lu, Chris Lu, Zoran Martinovic, Lindsay McCallum, Josh McGrath, Scott McKinney, Aidan McLaughlin, Song Mei, Steve Mostovoy, Tong Mu, Gideon Myles, Alexander Neitz, Alex Nichol, Jakub
  Pachocki, Alex Paino, Dana Palmie, Ashley Pantuliano, Giambattista Parascandolo, Jongsoo Park, Leher Pathak, Carolina Paz, Ludovic Peran, Dmitry Pimenov, Michelle Pokrass, Elizabeth Proehl, Huida Qiu, Gaby Raila, Filippo Raso, Hongyu Ren, Kimmy Richardson, David Robinson, Bob Rotsted, Hadi Salman, Suvansh Sanjeev, Max Schwarzer, D.~Sculley, Harshit Sikchi, Kendal Simon, Karan Singhal, Yang Song, Dane Stuckey, Zhiqing Sun, Philippe Tillet, Sam Toizer, Foivos Tsimpourlas, Nikhil Vyas, Eric Wallace, Xin Wang, Miles Wang, Olivia Watkins, Kevin Weil, Amy Wendling, Kevin Whinnery, Cedric Whitney, Hannah Wong, Lin Yang, Yu~Yang, Michihiro Yasunaga, Kristen Ying, Wojciech Zaremba, Wenting Zhan, Cyril Zhang, Brian Zhang, Eddie Zhang, and Shengjia Zhao.
\newblock gpt-oss-120b \& gpt-oss-20b model card, 2025.
\newblock URL \url{https://arxiv.org/abs/2508.10925}.

\bibitem[Team et~al.(2025)Team, Kamath, Ferret, Pathak, Vieillard, Merhej, Perrin, Matejovicova, Ramé, Rivière, Rouillard, Mesnard, Cideron, bastien Grill, Ramos, Yvinec, Casbon, Pot, Penchev, Liu, Visin, Kenealy, Beyer, Zhai, Tsitsulin, Busa-Fekete, Feng, Sachdeva, Coleman, Gao, Mustafa, Barr, Parisotto, Tian, Eyal, Cherry, Peter, Sinopalnikov, Bhupatiraju, Agarwal, Kazemi, Malkin, Kumar, Vilar, Brusilovsky, Luo, Steiner, Friesen, Sharma, Sharma, Gilady, Goedeckemeyer, Saade, Feng, Kolesnikov, Bendebury, Abdagic, Vadi, György, Pinto, Das, Bapna, Miech, Yang, Paterson, Shenoy, Chakrabarti, Piot, Wu, Shahriari, Petrini, Chen, Lan, Choquette-Choo, Carey, Brick, Deutsch, Eisenbud, Cattle, Cheng, Paparas, Sreepathihalli, Reid, Tran, Zelle, Noland, Huizenga, Kharitonov, Liu, Amirkhanyan, Cameron, Hashemi, Klimczak-Plucińska, Singh, Mehta, Lehri, Hazimeh, Ballantyne, Szpektor, Nardini, Pouget-Abadie, Chan, Stanton, Wieting, Lai, Orbay, Fernandez, Newlan, yeong Ji, Singh, Black, Yu, Hui, Vodrahalli, Greff, Qiu,
  Valentine, Coelho, Ritter, Hoffman, Watson, Chaturvedi, Moynihan, Ma, Babar, Noy, Byrd, Roy, Momchev, Chauhan, Sachdeva, Bunyan, Botarda, Caron, Rubenstein, Culliton, Schmid, Sessa, Xu, Stanczyk, Tafti, Shivanna, Wu, Pan, Rokni, Willoughby, Vallu, Mullins, Jerome, Smoot, Girgin, Iqbal, Reddy, Sheth, Põder, Bhatnagar, Panyam, Eiger, Zhang, Liu, Yacovone, Liechty, Kalra, Evci, Misra, Roseberry, Feinberg, Kolesnikov, Han, Kwon, Chen, Chow, Zhu, Wei, Egyed, Cotruta, Giang, Kirk, Rao, Black, Babar, Lo, Moreira, Martins, Sanseviero, Gonzalez, Gleicher, Warkentin, Mirrokni, Senter, Collins, Barral, Ghahramani, Hadsell, Matias, Sculley, Petrov, Fiedel, Shazeer, Vinyals, Dean, Hassabis, Kavukcuoglu, Farabet, Buchatskaya, Alayrac, Anil, Dmitry, Lepikhin, Borgeaud, Bachem, Joulin, Andreev, Hardin, Dadashi, and Hussenot]{gemmateam2025gemma3technicalreport}
Gemma Team, Aishwarya Kamath, Johan Ferret, Shreya Pathak, Nino Vieillard, Ramona Merhej, Sarah Perrin, Tatiana Matejovicova, Alexandre Ramé, Morgane Rivière, Louis Rouillard, Thomas Mesnard, Geoffrey Cideron, Jean bastien Grill, Sabela Ramos, Edouard Yvinec, Michelle Casbon, Etienne Pot, Ivo Penchev, Gaël Liu, Francesco Visin, Kathleen Kenealy, Lucas Beyer, Xiaohai Zhai, Anton Tsitsulin, Robert Busa-Fekete, Alex Feng, Noveen Sachdeva, Benjamin Coleman, Yi~Gao, Basil Mustafa, Iain Barr, Emilio Parisotto, David Tian, Matan Eyal, Colin Cherry, Jan-Thorsten Peter, Danila Sinopalnikov, Surya Bhupatiraju, Rishabh Agarwal, Mehran Kazemi, Dan Malkin, Ravin Kumar, David Vilar, Idan Brusilovsky, Jiaming Luo, Andreas Steiner, Abe Friesen, Abhanshu Sharma, Abheesht Sharma, Adi~Mayrav Gilady, Adrian Goedeckemeyer, Alaa Saade, Alex Feng, Alexander Kolesnikov, Alexei Bendebury, Alvin Abdagic, Amit Vadi, András György, André~Susano Pinto, Anil Das, Ankur Bapna, Antoine Miech, Antoine Yang, Antonia Paterson, Ashish
  Shenoy, Ayan Chakrabarti, Bilal Piot, Bo~Wu, Bobak Shahriari, Bryce Petrini, Charlie Chen, Charline~Le Lan, Christopher~A. Choquette-Choo, CJ~Carey, Cormac Brick, Daniel Deutsch, Danielle Eisenbud, Dee Cattle, Derek Cheng, Dimitris Paparas, Divyashree~Shivakumar Sreepathihalli, Doug Reid, Dustin Tran, Dustin Zelle, Eric Noland, Erwin Huizenga, Eugene Kharitonov, Frederick Liu, Gagik Amirkhanyan, Glenn Cameron, Hadi Hashemi, Hanna Klimczak-Plucińska, Harman Singh, Harsh Mehta, Harshal~Tushar Lehri, Hussein Hazimeh, Ian Ballantyne, Idan Szpektor, Ivan Nardini, Jean Pouget-Abadie, Jetha Chan, Joe Stanton, John Wieting, Jonathan Lai, Jordi Orbay, Joseph Fernandez, Josh Newlan, Ju~yeong Ji, Jyotinder Singh, Kat Black, Kathy Yu, Kevin Hui, Kiran Vodrahalli, Klaus Greff, Linhai Qiu, Marcella Valentine, Marina Coelho, Marvin Ritter, Matt Hoffman, Matthew Watson, Mayank Chaturvedi, Michael Moynihan, Min Ma, Nabila Babar, Natasha Noy, Nathan Byrd, Nick Roy, Nikola Momchev, Nilay Chauhan, Noveen Sachdeva, Oskar
  Bunyan, Pankil Botarda, Paul Caron, Paul~Kishan Rubenstein, Phil Culliton, Philipp Schmid, Pier~Giuseppe Sessa, Pingmei Xu, Piotr Stanczyk, Pouya Tafti, Rakesh Shivanna, Renjie Wu, Renke Pan, Reza Rokni, Rob Willoughby, Rohith Vallu, Ryan Mullins, Sammy Jerome, Sara Smoot, Sertan Girgin, Shariq Iqbal, Shashir Reddy, Shruti Sheth, Siim Põder, Sijal Bhatnagar, Sindhu~Raghuram Panyam, Sivan Eiger, Susan Zhang, Tianqi Liu, Trevor Yacovone, Tyler Liechty, Uday Kalra, Utku Evci, Vedant Misra, Vincent Roseberry, Vlad Feinberg, Vlad Kolesnikov, Woohyun Han, Woosuk Kwon, Xi~Chen, Yinlam Chow, Yuvein Zhu, Zichuan Wei, Zoltan Egyed, Victor Cotruta, Minh Giang, Phoebe Kirk, Anand Rao, Kat Black, Nabila Babar, Jessica Lo, Erica Moreira, Luiz~Gustavo Martins, Omar Sanseviero, Lucas Gonzalez, Zach Gleicher, Tris Warkentin, Vahab Mirrokni, Evan Senter, Eli Collins, Joelle Barral, Zoubin Ghahramani, Raia Hadsell, Yossi Matias, D.~Sculley, Slav Petrov, Noah Fiedel, Noam Shazeer, Oriol Vinyals, Jeff Dean, Demis Hassabis,
  Koray Kavukcuoglu, Clement Farabet, Elena Buchatskaya, Jean-Baptiste Alayrac, Rohan Anil, Dmitry, Lepikhin, Sebastian Borgeaud, Olivier Bachem, Armand Joulin, Alek Andreev, Cassidy Hardin, Robert Dadashi, and Léonard Hussenot.
\newblock Gemma 3 technical report, 2025.
\newblock URL \url{https://arxiv.org/abs/2503.19786}.

\bibitem[Tsai and Bagdasarian(2025)]{tsai2025contextualagentsecuritypolicy}
Lillian Tsai and Eugene Bagdasarian.
\newblock Contextual agent security: A policy for every purpose, 2025.
\newblock URL \url{https://arxiv.org/abs/2501.17070}.

\bibitem[Wang et~al.(2025)Wang, Ma, Wang, Wu, Ji, Chen, Li, and Yuan]{wang2025surveyllmbasedagentsmedicine}
Wenxuan Wang, Zizhan Ma, Zheng Wang, Chenghan Wu, Jiaming Ji, Wenting Chen, Xiang Li, and Yixuan Yuan.
\newblock A survey of llm-based agents in medicine: How far are we from baymax?, 2025.
\newblock URL \url{https://arxiv.org/abs/2502.11211}.

\bibitem[Yang et~al.(2025)Yang, Li, Yang, Zhang, Hui, Zheng, Yu, Gao, Huang, Lv, Zheng, Liu, Zhou, Huang, Hu, Ge, Wei, Lin, Tang, Yang, Tu, Zhang, Yang, Yang, Zhou, Zhou, Lin, Dang, Bao, Yang, Yu, Deng, Li, Xue, Li, Zhang, Wang, Zhu, Men, Gao, Liu, Luo, Li, Tang, Yin, Ren, Wang, Zhang, Ren, Fan, Su, Zhang, Zhang, Wan, Liu, Wang, Cui, Zhang, Zhou, and Qiu]{yang2025qwen3technicalreport}
An~Yang, Anfeng Li, Baosong Yang, Beichen Zhang, Binyuan Hui, Bo~Zheng, Bowen Yu, Chang Gao, Chengen Huang, Chenxu Lv, Chujie Zheng, Dayiheng Liu, Fan Zhou, Fei Huang, Feng Hu, Hao Ge, Haoran Wei, Huan Lin, Jialong Tang, Jian Yang, Jianhong Tu, Jianwei Zhang, Jianxin Yang, Jiaxi Yang, Jing Zhou, Jingren Zhou, Junyang Lin, Kai Dang, Keqin Bao, Kexin Yang, Le~Yu, Lianghao Deng, Mei Li, Mingfeng Xue, Mingze Li, Pei Zhang, Peng Wang, Qin Zhu, Rui Men, Ruize Gao, Shixuan Liu, Shuang Luo, Tianhao Li, Tianyi Tang, Wenbiao Yin, Xingzhang Ren, Xinyu Wang, Xinyu Zhang, Xuancheng Ren, Yang Fan, Yang Su, Yichang Zhang, Yinger Zhang, Yu~Wan, Yuqiong Liu, Zekun Wang, Zeyu Cui, Zhenru Zhang, Zhipeng Zhou, and Zihan Qiu.
\newblock Qwen3 technical report, 2025.
\newblock URL \url{https://arxiv.org/abs/2505.09388}.

\bibitem[Zhang et~al.(2025)Zhang, Su, Chen, Bertino, Zhang, and Li]{zhang2025llmagentsemploysecurity}
Kaiyuan Zhang, Zian Su, Pin-Yu Chen, Elisa Bertino, Xiangyu Zhang, and Ninghui Li.
\newblock Llm agents should employ security principles, 2025.
\newblock URL \url{https://arxiv.org/abs/2505.24019}.

\end{thebibliography}

\appendix

\section{Technical Appendices and Supplementary Material}

\subsection{POLICY-TREE-GEN PROMPTS}

\vspace{1em}
\begin{figure}[ht]
  \centering
\begin{tcolorbox}[
  colback=gray!10,
  colframe=teal!70!black,
  title=\textbf{\large LLM Prompt: Pass 1 — Parse Document \& Classify Security Instructions}
]
{\large

\textbf{Role \& Goal}

You are a document analysis expert specializing in data security for LLM-powered applications. Read a document describing an LLM-based system, infer its logical structure, and extract only those instructions relevant to data and guardrail security.

\medskip
\textbf{Important:} Ignore all specific examples in this step. Focus strictly on the \emph{instructions} themselves.

\bigskip
\textbf{Input}

<Document text>

\medskip
\textbf{Task}

Return a single JSON object representing the document's hierarchy. For each extracted instruction, you \emph{must} include:
\begin{itemize}
  \item The exact, verbatim \texttt{source\_span} copied from the document.
  \item A \texttt{category} chosen from 4 CATEGORIES.
\end{itemize}

\medskip
\textbf{ID Categories \& Guidance}

\begin{itemize}
  \item \texttt{ID-I}: Defines what the system should accept (in-domain, topical, acceptable inputs).  
  \emph{Exclude:} outputs, rejection rules, metadata, formatting, or negative conditions.
  
  \item \texttt{ID-O}: Defines correct system responses (including handling invalid/unsupported/out-of-domain inputs).  
  \emph{Exclude:} malformed prompts, format/schema rules, valid inputs, or outputs.
  
  \item \texttt{OOD-I}: Defines what the system must reject (invalid, unsupported, or out-of-domain inputs).  
  \emph{Include:} logic/guardrails for blocking inputs.  
  \emph{Exclude:} formatting, valid inputs, or outputs.
  
  \item \texttt{OOD-O}: Defines incorrect or forbidden system responses (hallucinations, policy violations, leaking PII, misleading answers).  
  \emph{Include:} logic/guardrails for blocking outputs.  
  \emph{Exclude:} formats, constraints, valid outputs, or inputs.
\end{itemize}

\textbf{Exclusion Rule:} If an instruction does not clearly fit one of the valid categories above, \emph{omit it} from the output.

\medskip
\textbf{Required JSON Shape}

Root key: \texttt{"document\_structure"} (list).  
Each element:
\begin{itemize}
  \item \texttt{"topic"}: Heading
  \item \texttt{"content"}: Instructions with \texttt{source\_span}, \texttt{category}, \texttt{sub\_instructions}
  \item \texttt{"children"}: Sub-topics (same format)
\end{itemize}

}
\end{tcolorbox}

\caption{Prompt for hierarchical parsing in \texttt{POLICY-TREE-GEN}}
\label{fig:llm-pass1-parse-classify}
\end{figure}

\clearpage

\vspace{1em}
\begin{figure}[ht]
  \centering
\begin{tcolorbox}[
  colback=gray!10,
  colframe=teal!70!black,
  title=\textbf{\large LLM Prompt: Pass 2 — Extract \& Classify Policy Examples}
]
{\large

\textbf{Role \& Goal}

You are a policy expert. For a given policy instruction, find \emph{all relevant examples} that already exist in the provided document. \textbf{Do not} invent or generate new examples.

\bigskip
\textbf{Inputs}

\begin{description}[style=nextline,leftmargin=1em]
  \item[\texttt{POLICY INSTRUCTION}:] 
\begin{verbatim}
"{instruction_text}"
\end{verbatim}

\begin{verbatim}
---
{document_text}
---
\end{verbatim}
\end{description}

\bigskip
\textbf{Task}

Return a single JSON object with a top-level key \texttt{"examples"} containing a list of example objects found in the document that are relevant to the policy instruction.

\bigskip
\textbf{Constraints}

\begin{itemize}
  \item \textbf{No generation}: Only extract examples that \emph{already} exist in the document.
  \item Include every relevant example; omit non-relevant content.
\end{itemize}

\bigskip
\textbf{Example Object Schema}

Each example object must include:
\begin{itemize}
  \item \texttt{"example\_text"} — The text of the example.
  \item \texttt{"source\_span"} — The exact, verbatim sentence copied from the document.
  \item \texttt{"polarity"} — Positive or Negative: 
    \begin{itemize}
      \item \texttt{positive}: the example follows the instruction.
      \item \texttt{negative}: the example is a counter-example.
    \end{itemize}
\end{itemize}

\bigskip
\textbf{Empty Result Handling}

If no relevant examples are found, return:
\begin{verbatim}
{"examples": []}
\end{verbatim}

\bigskip
\textbf{In-Domain Example (context only)}


}
\end{tcolorbox}

\caption{Prompt for extracting existing policy examples in \texttt{POLICY-TREE-GEN}}
\label{fig:policy-examples-pass2}
\end{figure}

\subsection{POLICY-AS-PROMPT OUTPUT}

\vspace{1em}
\begin{figure}[ht]
  \centering
\begin{tcolorbox}[
    colback=gray!10,
    colframe=teal!70!black,
    title=\textbf{\large LLM Prompt: Policy Compliance Classification Task}
  ]
  {\large
  
You are a meticulous security and policy compliance analyst. Your sole responsibility is to analyze a user’s input and determine if it complies with a strict set of predefined rules.

\medskip
\textbf{Policy Document and Classification Reference Data:}\\

\begin{description}[leftmargin=1em, style=nextline]
  \item[Rules for ID Inputs]
  \begin{itemize}
    \item It is acceptable to automate the analysis of HR support tickets only if the content is free of personal data.
  \end{itemize}

  \textbf{Example of a ID anonymized HR support ticket:}  
  \begin{quote}
    "Request: I am unable to access the HR portal to update my benefits information.  
    Error: System shows 'Access Denied' despite using correct login.  
    Resolution attempted: Cleared cache, tried on a different browser.  
    Request: Please escalate to IT for resolution."  
  \end{quote}

  \item[Rules for OOD Inputs]  
  \begin{itemize}
    \item Including any personal data in AI analysis or outputs without prior automated deletion or anonymization is a direct violation of the policy.
    \item Examples of personal data that must be removed include: names, email addresses, employee IDs, government IDs, birth dates, home addresses, salary details, beneficiary information, etc.
  \end{itemize}

  \textbf{Example of an OOD non-anonymized HR support ticket:}  
  \begin{quote}
    "Request: John Smith (Employee ID: 45832) cannot access his HR portal.  
    Email: john.smith@company.com  
    Issue: Unable to update dependent details for his son born on 12/03/2020.  
    Request: Please resolve urgently."  
  \end{quote}

  \textbf{Example of an OOD non-HR ticket (out of scope):}  
  \begin{quote}
    "Request: The office coffee machine on the 3rd floor is broken again.  
    Issue: Display shows 'Water Pump Error.'  
    Request: Please arrange maintenance."  
  \end{quote}
\end{description}


\textbf{Output Format:}

Return a single JSON object with the following keys:
\begin{itemize}
  \item \texttt{"classification"}: Either \texttt{"ID"} or \texttt{"OOD"}.
  \item \texttt{"reasoning"}: A one-sentence explanation referencing the rule followed or violated.
\end{itemize}

\bigskip

\textbf{Task:}

Analyze the following \texttt{DOCUMENT} and provide your classification:

\medskip

\noindent\hspace{-1em}

\bigskip

\textbf{Your JSON response:}
}
\end{tcolorbox}

\caption{Policy-As-Prompt for HR Application Input Classifier}
  \label{fig:policy-classification}
\end{figure}

\clearpage

\vspace{0.75em}
\begin{figure}[ht]
  \centering
\begin{tcolorbox}[
  colback=gray!10,
  colframe=teal!70!black,
  title=\textbf{\large LLM Prompt: Policy Compliance Classification (Content-Focused)}
]
{\large

You are a security/policy analyst. Analyze the user’s query and decide if the \textbf{content of the INPUT} is allowed under the policy. Ignore tooling/functions; focus only on what data the INPUT contains and what the OUTPUT must contain.

\medskip
\textbf{Policy — Allowed vs. Disallowed Content}

\begin{description}[leftmargin=1em, style=nextline]

  \item[ALLOWED INPUT CONTENT (no personal data, no secrets)]
  \begin{itemize}
    \item \emph{Alert metadata:} alert/case IDs, timestamps, severity, source system, rule/signature names, tags.
    \item \emph{Technical indicators (IoCs):} file hashes, public IPs, domains, URLs (no credentials), file/process names, registry keys, event codes.
    \item \emph{System/asset context:} asset IDs, generic hostnames (not derived from real names), service/application names, environment (prod/stage/dev), non-personal ticket/channel IDs.
    \item \emph{Log details (sanitized):} network flow summaries, HTTP metadata (without tokens/cookies), counts, time windows, error codes; fields may be pseudonymized or hashed.
    \item \emph{Org context (non-personal):} role or team names (e.g., “SRE on-call”), runbook references, anonymized user placeholders (\texttt{user\_A}, hashed IDs).
    \item \emph{Threat intel:} malware family names, campaign/actor names, rule IDs, confidence scores.
    \item \emph{Attachments/screenshots} only if redacted/pseudonymized so no personal data or secrets are visible.
  \end{itemize}

  \item[DISALLOWED INPUT CONTENT]
  \begin{itemize}
    \item \textbf{Personal data (PII):} names, emails, phone numbers, home addresses, government IDs, birth dates, employee/customer IDs tied to individuals.
    \item \textbf{Sensitive records:} health, financial, or HR records; salary/beneficiary details; customer lists; resumes.
    \item \textbf{Secrets/credentials:} passwords, API keys, private keys, tokens, session IDs, cookies, JWTs, MFA codes, OAuth headers.
    \item \textbf{Raw dumps containing PII/secrets:} browser/localStorage, memory dumps, unredacted chat/email transcripts, authorization headers.
    \item \textbf{Identifiers revealing a person:} hostnames/usernames or filenames that embed real names (must be pseudonymized before submission).
    \item \textbf{Out-of-scope facility issues} (e.g., coffee machine maintenance).
  \end{itemize}

\end{description}

\textbf{Output Contract}

Return a single JSON object:
\begin{itemize}
  \item \texttt{"classification"}: \texttt{"ID"} if the INPUT only contains \emph{Allowed} content; otherwise \texttt{"OOD"}.
  \item \texttt{"reasoning"}: One concise sentence citing the specific rule followed/violated.
\end{itemize}
\bigskip
\textbf{Task:} Analyze the following \texttt{DOCUMENT} and classify it:

\medskip
\noindent\hspace{-1em}

\bigskip
\textbf{Your JSON response:}
}
\end{tcolorbox}

\caption{Content-First Policy-As-Prompt for Security Input Classification}
\label{fig:policy-classification-2}
\end{figure}

\clearpage

\subsection{Evaluation for POLICY-TREE-GEN}

\begin{table}[htbp]
\caption{Evaluation Metrics for \texttt{POLICY-TREE-GEN}}
\label{tab:perc_hr_secops_less_useful}
\centering
\scriptsize
\begin{tabular}{llrrrrrr}
\toprule
\textbf{Application} & \textbf{Model} & \textbf{Det P (\%)} & \textbf{Micro-F1 (\%)} & \textbf{Span Exact (\%)} & \textbf{Token-F1} & \textbf{Substr (\%)} & \textbf{Emb Cos} \\
\midrule
HR      & O1               & 68.6 & 30.0 & 91.7 & 0.999 & 100.0 & 0.996 \\
HR      & GPT-OSS 120B     & 42.1 & 9.4  & 0.0  & 0.910 & 87.5  & 0.905 \\
HR      & Llama 405B       & 40.0 & 10.9 & 75.0 & 1.000 & 100.0 & 0.973 \\
HR      & Claude 3.5       & 40.0 & 4.0  & 50.0 & 0.992 & 100.0 & 0.986 \\
\midrule
SOC  & O1               & 26.9 & 22.6 & 85.7 & 0.987   & 100   & 0.958   \\
SOC  & GPT-OSS 120B     & 66.7 & 10.3 & 0.0  & 0.974 & 0.0   & 0.881 \\
SOC  & Llama 405B       & 40.0 & 9.8  & 0.0  & 0.842 & 100.0 & 0.790 \\
SOC  & Claude 3.5       & 100.0& 5.4  & 0.0  & 0.818 & 100.0 & 0.782 \\
\bottomrule
\end{tabular}
\end{table}

\end{document}